# Communication Modalities for Supervised Teleoperation in Highly Dexterous Tasks – Does one size fit all?

Tian Zhou, Maria E. Cabrera, *Student Member, IEEE*, and Juan P. Wachs, *Member, IEEE*

*Abstract*— This study tries to explain the connection between communication modalities and levels of supervision in teleoperation during a dexterous task, like surgery. This concept is applied to two surgical related tasks: incision and peg transfer. It was found that as the complexity of the task escalates, the combination linking human supervision with a more expressive modality shows better performance than other combinations of modalities and control. More specifically, in the peg transfer task, the combination of speech modality and action level supervision achieves shorter task completion time (77.1 ±3.4 s) with fewer mistakes (0.20±0.17 pegs dropped).

## I. INTRODUCTION

Teleoperation is an area of paramount importance in modern robotics due to the large number of applications where it is being utilized, such as search and rescue and healthcare [1], [2]. Teleoperation enables operators to control a robot so it can perform a task, in a friendly or hostile environment. While the teleoperation scheme can help to remove the human operator from danger, it was found that it does not necessarily increase efficiency and productivity [3].

Since the fundamental principle of teleoperation builds on the human operator remotely controlling the robot, the overall performance is affected by the limited perception and situation awareness of the operator [4]. In addition, there is a cognitive toll on the operator due to the continuous monitoring and control of the robot. This cognitive burden may be reduced by allowing the robot to undertake some routines [3]. On the other hand, a too high reliance on the robot may be risky since the "human in the loop" is required for decision-making in situations the robot is not capable to resolve [5].

Given that different interaction modalities are associated with different channels' capacity, bandwidth and degrees of freedom (Shannon–Hartley theorems [6]), it makes sense to associate the levels of supervision with modalities capable of expressing the level of complexity. There is no systematic study presented so far to explore the suitability of the communication modalities mapped to levels of autonomy.

This study tries to fill this gap by understanding the connection between the type of modality used to communicate with a robot, and the level of autonomy during task completion. The selected case study is based on two surgically related tasks: incision and peg-pole transfer. In the scope of robotic assisted surgery (RAS), supervised autonomy can be incorporated in sections of the task (surgery) which are constant, repetitive, and do not require as much involvement of the operator.

## II. SYSTEM OVERVIEW

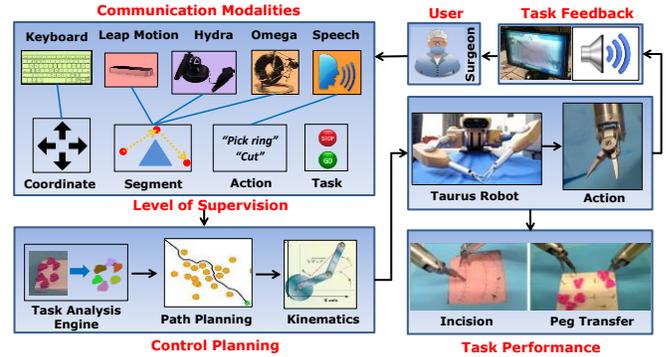

Figure 1. System Overview

The implementation of the system is based on the Taurus robot which can make incisions and transfer parts with its two 7-DOF independently controlled manipulators [7]. The system architecture is presented in Figure 1. There are four levels of supervision implemented: (i) coordinate level where the instructions are in the form of basic motion increment or decrement in each DOF; (ii) segment level which consists segment motions where streams of landmarks are sent to the robot as via points to follow; (iii) action level where primitive commands encompass a larger set of robotic movements to accomplish the intended action; (iv) task level in which the operator is disengaged from the robot completely and the robot needs to rely on autonomy to finish the task. Five different communication modalities were incorporated into the system, ranging from traditional interfaces like keyboard, to new motion sensing interfaces like Leap Motion, Hydra and Omega, and lastly to more natural interaction channels like speech, using Sphnix library [8].

The heuristic followed consists of associating each level of supervision with a modality capable to express the corresponding complexity of the given command. Let $m_i$ be the modality of communication, so that $m_i \in M$ (the set of all the N communication modalities). Let $s_j$ be the level of supervised autonomy, so that $s_j \in S$ (the set of all the Q supervision levels). Let the function $r(*)$ be a function that returns the ranking of a modality or supervision level. The ranking value represents the place in an ordered list according to the value of an objective or subjective function (e.g. channel capacity, or level of complexity or compositionality). The higher the objective function value is, the lower the ranking value is. The best ranking is 1, and the worst is N and Q respectively. Then, a heuristic can be formulated in the following:

$$\xi = \{(m_i, s_j) \in M \times S : r(m_i) = r(s_j)\}, 1 \leq i \leq N; 1 \leq j \leq Q \quad (1)$$

In other words, the modalities and levels of supervision are mapped so their rankings match. Table 1 shows the ranking of the levels of supervision available, as well as the matching communication modalities after applying the heuristic.

TABLE 1 RANKING FOR MAPPING LEVEL OF SUPERVISION TO COMMUNICATION MODALITIES

| Levels of Supervision | Communication Modalities |
|---|---|
| 1. Task Level | 1. Autonomous (No modality) |
| 2. Action Level | 2. Speech |
| 3. Segment Level | 3. Leap Motion, Hydra, Omega |
| 4. Coordinate Level | 4. Keyboard |

Given a certain level of supervision, the task analysis engine will expand from simple point interpolation to full autonomous mode. After recognizing the user's commands, a motion trajectory is generated based on path planning and robotic kinematics. As the robot executes the commands, feedback is sent back to the user visually in the form of 3D images. Stiffness, position and orientation of the tooltips are presented to the user as feedback through color and sound cues throughout the task.

## III. EXPERIMENTS

Two tasks were selected from the Fundamentals of Laparoscopic Surgery (FLS) [9], namely incision and peg transfer. The first task is to complete a horizontal incision using a scalpel mounted on the robot's end-effector, while the peg transfer task is a reduced version moving only 3 rings from one side of a pegboard to the other. In each experiment, various metrics were used to measure the performance of the different combinations between interfaces and levels of supervision. Ten engineering students were recruited, and each was assigned to command Taurus using two out of the five available interfaces through the task, for five repetitions each modality. The order of the two modalities was randomized to compensate for the learning of the task.

Figure 2 shows the average error in distance from the target trajectory (left), and the average root mean square (RMS) of the depth trajectory (right), with different letters indicating groups with significant difference ($p<0.001$ for Welch's ANOVA test and for Games-Howell post hoc test). Computer controlled modalities (Group A) showed the least distance errors and depth fluctuation among all groups.

The metrics regarding the peg transfer experiment are shown in Table 2, with letters indicating different groups with statistical significance. There is a trend of increment in time and decrement of peg drops when the level of supervision increases. As the user is more engaged, the task takes more time to be finished with higher accuracy and fewer mistakes.

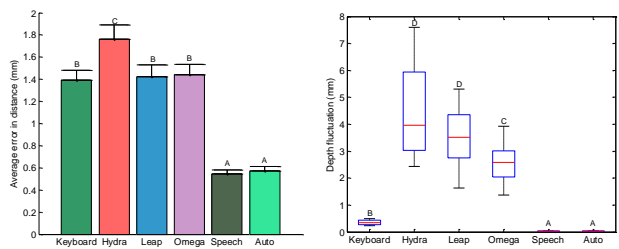

Figure 2. (left) Average distance error. (right) Boxplot of depth fluctuation.

TABLE 2. AVERAGE PERFORMANCE FOR THE PEG TRANSFER TASK

| Interface | Time ±CI (s) | Peg Drops ±CI | Learning Rate |
|---|---|---|---|
| Leap Motion | $360.2^D$ ±71.0 | $1.30^C$ ±0.62 | **69.86%** |
| Keyboard | $271.9^D$ ±37.3 | **$0.15^A$** ±0.16 | 93.83% |
| Omega | $121.3^C$ ±13.3 | $0.20^A$ ±0.22 | 85.69% |
| Hydra | $107.9^C$ ±10.0 | $0.20^A$ ±0.22 | 92.71% |
| Speech | $77.1^B$ ±3.4 | $0.20^A$ ±0.17 | 97.65% |
| Auto | **$60.6^A$** ±1.9 | $0.70^B$ ±0.28 | 96.85% |

## IV. CONCLUSION

This work presents a study of the relationship between communication modalities and levels of supervision. A heuristic mapping was established between the two concepts and two surgical experiments were conducted to test the performance of the selected mapping.

It was found that relatively simple tasks (such as horizontal incision) can be solved optimally relying on robot autonomy and minimal human supervision. On the other hand, when the complexity of the task increases, modalities that engage the user more, providing higher level of supervision, show increasing performance as it complements the robot's limited sensing and computational capabilities with the human, at the cost of longer task completion time.

## ACKNOWLEDGMENT

This publication was made possible by the NPRP award (NPRP 6-449-2-181) from the Qatar National Research Fund (a member of The Qatar Foundation). The statements made herein are solely the responsibility of the authors.

## REFERENCES

[1] A. M. Okamura, M. J. Mataric, and H. I. Christensen, "Medical and health-care robotics," *Robot. Autom. Mag.*, vol. 17, no. 3, pp. 26–27, 2010.
[2] M. Baker, R. Casey, B. Keyes, and H. A. Yanco, "Improved interfaces for human-robot interaction in urban search and rescue.," in *SMC (3)*, 2004, pp. 2960–2965.
[3] G. Cheng and A. Zelinsky, "Supervised autonomy: A paradigm for teleoperating mobile robots," in *Intelligent Robots and Systems, 1997. IROS'97., Proceedings of the 1997 IEEE/RSJ International Conference on*, 1997, vol. 2, pp. 1169–1176.
[4] D. B. Kaber, E. Onal, and M. R. Endsley, "Design of automation for telerobots and the effect on performance, operator situation awareness, and subjective workload," *Hum. Factors Ergon. Manuf.*, vol. 10, no. 4, pp. 409–430, 2000.
[5] G. Cheng and A. Zelinsky, "Supervised autonomy: A framework for human-robot systems development," *Auton. Robots*, vol. 10, no. 3, pp. 251–266, 2001.
[6] C. E. Shannon, "A Mathematical Theory of Communication," *SIGMOBILE Mob Comput Commun Rev*, vol. 5, no. 1, pp. 3–55, Jan. 2001.
[7] T. Zhou, M. E. Cabrera, and J. P. Wachs, "Touchless telerobotic surgery – is it possible at all?," in *The Twenty-Ninth AAAI Conference on Artificial Intelligence (accepted)*, 2015.
[8] W. Walker, P. Lamere, P. Kwok, B. Raj, R. Singh, E. Gouvea, P. Wolf, and J. Woelfel, "Sphinx-4: A flexible open source framework for speech recognition," 2004.
[9] K. M. Augestad, T. Chomutare, J. G. Bellika, A. Budrionis, R.-O. Lindsetmo, C. P. Delaney, and M. M. M. (M3) P. Group*, "Clinical and Educational Benefits of Surgical Telementoring," in *Simulation Training in Laparoscopy and Robotic Surgery*, H. R. H. Patel and J. V. Joseph, Eds. Springer London, 2012, pp. 75–89.